# A phenotype-driven and evidence-governed framework for knowledge graph enrichment and hypotheses discovery in population data

Adela BÂRA, Simona-Vasilica OPREA
*Department of Economic Informatics and Cybernetics, Bucharest University of Economic Studies, Bucharest, Romania*
*Corresponding author. E-mail: simona.oprea@csie.ase.ro

**Abstract:** Current knowledge graph (KG) construction methods are confirmatory, focusing on recovering known relationships rather than identifying novel or context-dependent nodes. This paper proposes a phenotype-driven and evidence-governed framework that shifts the paradigm toward structured hypothesis discovery and controlled KG expansion. The approach integrates graph neural networks (GNNs) for phenotype discovery, causal inference, probabilistic reasoning and large language models (LLMs) for hypothesis generation and claim extraction within a unified pipeline. The framework prioritizes relationships that are both structurally supported by data and underexplored in the literature. KG expansion is formulated as a multi-objective optimization problem, where candidate claims are jointly evaluated in terms of relevance, structural validation and novelty. Pareto-optimal selection enables the identification of non-dominated claims that balance confirmation and discovery, avoiding trivial or redundant knowledge inclusion. Experiments on heterogeneous population datasets demonstrate that the proposed framework produces more interpretable phenotypes, reveals context-dependent causal structures and generates high-quality claims that align with both data and scientific evidence. Compared to rule-based and LLM-only baselines, the method achieves the best trade-off across plausibility, novelty, validation and relevance. In retrieval-augmented settings, it significantly improves performance (Recall@5=0.98) while reducing hallucination rates (0.05), highlighting its effectiveness in grounding LLM outputs.

## 1. Introduction

The progress in data-driven modeling, particularly in graph-based learning [1], causal discovery [2], [3] and large language models (LLMs) [4], has significantly improved the ability to explain complex systems. Most existing approaches, however, remain predominantly *confirmatory*: they aim to identify patterns, validate known relationships or predict outcomes based on observed data. Such approaches are inherently limited in their ability to systematically uncover what is missing, underexplored or weakly supported in current knowledge. In parallel, several research directions have emerged that partially address this limitation: (a) Knowledge graph (KG) [5] completion methods attempt to infer missing edges based on structural patterns, while causal discovery frameworks identify plausible dependencies from data; (b) Further, LLM-based systems have been used to generate hypotheses and research questions from unstructured knowledge sources. However, these paradigms are typically developed in isolation and lack a unified mechanism to jointly reason about data-driven structure, probabilistic dependencies and existing scientific evidence [6], [7]. This paper addresses this gap by proposing a phenotype-driven, evidence-governed KG enrichment framework that explicitly shifts the modeling paradigm from *"explaining what is known"* to *"suggesting what might be missing, plausible and worth investigating"*.

The main objective of the paper is to design a knowledge-based modeling and expansion framework using data-driven and plausible hypotheses related to causal and probabilistic inference of existing relationships in population health studies. The population data (user states) is modeled as graph nodes with cross-domain interactions (for e.g., psychological, nutrition, lifestyle, medication) and clustered into phenotypes using graph neural networks (GNNs) to identify latent subpopulations with shared structural patterns. A soft-matching approach is used to assign each user state to multiple phenotypes and to leverage phenotype-specific representations as an intermediate abstraction layer that captures heterogeneous, cross-domain and context-dependent system states. Rather than assuming global relationships across the population, the graph models localized structures within phenotypes. For each phenotype, causal discovery using NOTEARS [8] is applied to infer phenotype-specific directional dependencies, and probabilistic modeling using Bayesian Networks [9] is used to quantify uncertainty and conditional influence in the clusters.

Each cluster is modeled as a phenotype state (PS) and used by a LLM to generate causal and probabilistic-driven hypotheses related to the influence of predictors and mediators over the selected outcomes (for e.g., *"Does social engagement intervention improve depressive symptoms in high stressed students' population"*). The hypotheses are transformed into structured queries, and, for each hypothesis, the most relevant papers are retrieved from PubMed along with their claims and recommendations. A structured mechanism for identifying underexplored and latent relationships for graph expansion is identified that instead of focusing solely on strong or statistically significant influences, it analyzes weak but structurally valid dependencies, indirect or multi-hop relationships, cross-phenotype variability with low-support relationships in the literature. These signals are combined into a novelty-plausibility scoring (NPS) function, enabling the prioritization of hypotheses that are both data-consistent and insufficiently explored in existing research. Based on it, the framework formulates knowledge graph expansion as a multi-objective optimization problem, where each candidate claim is evaluated jointly in terms of relevance, validation and novelty–plausibility score (NPS). Rather than imposing a single aggregate ranking, the method identifies a set of Pareto-optimal claims, representing non-dominated solutions that balance these competing objectives. This allows the selection of claims that are not only strongly supported by evidence and consistent with the underlying graph structure but also include structurally plausible yet underexplored relationships. Thus, the KG is extended with the scientific findings that are used to ground the health recommendations and further explore novel and plausible hypotheses.

To guide the development and evaluation of the proposed framework, the study addresses the following research questions:

*RQ1*: How can heterogeneous user states be transformed into interpretable phenotype representations that capture cross-domain interactions, feature-level and structural dependencies?

*RQ2*: To what extent does phenotype-specific modeling (via GNN clustering, causal discovery and probabilistic inference) improve the identification of context-dependent relationships compared to global approaches?

*RQ3*: How can LLMs be constrained by causal and probabilistic structures to generate scientifically plausible and non-trivial hypotheses?

*RQ4*: How can we systematically identify underexplored or missing relationships by combining structural, probabilistic and literature-based signals? Can the proposed framework effectively support KG enrichment and user-centric recommendations by integrating validated claims while maintaining consistency with both data and prior knowledge?

*RQ5*: How robust is the framework in handling new or unseen user states, including soft phenotype matching and anomaly-driven discovery of new patterns?

This paper makes the following key contributions:

(1) A unified phenotype data-driven framework that combines causal discovery, probabilistic inference and LLM-based reasoning for hypotheses generation;

(2) An evidence-governed graph enrichment strategy, where KG expansion is controlled through joint validation using causal discovery, probabilistic inference and literature-derived claims;

(3) A multi-objective optimization mechanism for knowledge graph expansion, where candidate claims are evaluated based on relevance, structural validation and novelty–plausibility and Pareto-optimal selection is used to identify non-dominated claims that balance confirmation and discovery;

(4) A dynamic adaptation mechanism for new user states, supporting soft phenotype assignment, anomaly detection and incremental discovery of new phenotypes.

Thus, the proposed framework doesn't expand the graph to store common facts, but it expands it selectively to ground the uncertain, less visible, phenotype-relevant knowledge that matters most for hallucination reduction and tailored recommendations. This contribution is particularly important in the context of retrieval-augmented generation (RAG). LLMs already encode common and frequently observed relationships with high confidence and exhibit relatively low hallucination rates for such cases. In contrast, underexplored or context-specific relationships are more prone to hallucination. The proposed framework focuses graph enrichment on regions of the knowledge space where explicit grounding is most beneficial, thereby improving reliability and supporting more tailored recommendations.

## 2. Literature review

Early foundational work highlights their ability to model complex real entities and relationships using graph-based representations, supported by specialized query and validation languages, as well as hybrid reasoning mechanisms combining symbolic and data-driven approaches [10]. Subsequent surveys further consolidate the role of KGs in artificial intelligence (AI), emphasizing their capacity to organize vast knowledge repositories and enable advanced reasoning capabilities across domains [11], [12], [13].

A broader systematic perspective is provided in [14], where KGs are positioned as a central component of AI systems. It outlines their dual role: enabling intelligent applications and supporting diverse real use cases. At the same time, it identifies critical technical challenges, including knowledge acquisition, embedding learning, graph completion, fusion of heterogeneous sources and reasoning under uncertainty. Traditionally built through expert curation, these graphs are increasingly generated and enhanced using automated and machine learning (ML) methods [15]. Representation learning techniques allow these graphs to be transformed into lower-dimensional forms, enabling predictive tasks across genomics, drug discovery and clinical research.

The convergence of KGs and LLMs represents a rapidly evolving research direction aimed at addressing limitations such as hallucinations and lack of factual grounding. The KG-based Thought (KGT) framework proposed in [16] exemplifies this integration by enhancing LLM outputs with structured knowledge, significantly improving factual accuracy in biomedical question answering tasks. Similarly, hybrid architectures combining KGs and LLMs have demonstrated strong performance in mitigating misinformation and improving reasoning reliability [17].

Visualization and interaction frameworks such as KNOWNET further extend this paradigm by aligning LLM-generated outputs with verified KG evidence, enabling structured exploration and iterative knowledge discovery [18]. In addition, LLM-augmented KG construction approaches have been proposed to automate ontology design and relation extraction, demonstrating strong performance across domains such as disaster resilience modelling [19] and disease-specific graph construction [20].

Other studies highlight the role of KGs in enhancing LLM interpretability and consistency. For instance, integrating structured graph representations into LLM pipelines improves query answering over clinical data and reduces inconsistencies in generated responses [21]. Moreover, advanced predictive frameworks such as LLM-DG leverage KG-like representations to model both inter- and intra-patient relationships, significantly improving diagnostic prediction performance [22]. Additionally, explainability and trust are critical in healthcare AI systems, and KGs play a central role in addressing these concerns. By explicitly modeling entities and their relationships, KGs provide transparent reasoning paths that can be interpreted by domain experts. A comprehensive review in [23] demonstrates how KGs enhance explainability across different stages of AI pipelines-before, during and after model training-supporting applications such as adverse drug reaction detection, misinformation identification and clinical decision support.

Despite advances in constructing biomedical KGs, tools for their exploration and visualization remain limited. The KGEV framework addresses this gap by providing an interactive platform for building, querying and analyzing KGs [24]. A significant body of research focuses on constructing KGs from Electronic Health Records (EHRs) and real clinical data. EHR-based KGs provide rich, patient-centric representations that support downstream tasks such as diagnosis prediction, clinical recommendations and knowledge inference [25]. However, challenges such as data heterogeneity, high dimensionality and dynamic updates remain critical issues. Advanced frameworks such as ARCH [26] address these challenges by integrating structured and unstructured EHR data into unified graph representations, achieving strong performance in tasks like disease phenotyping and drug side-effect detection. Similarly, clinical KG construction pipelines leveraging natural language processing (NLP) tools enable the transformation of unstructured clinical notes into structured knowledge suitable for predictive analytics [27]. Other approaches emphasize fully automated KG curation using ontology-driven and ML techniques, enabling predictive modeling for complex diseases [28]. Large-scale resources further demonstrate the potential of real data by extracting millions of disease-phenotype associations from clinical narratives [29].

To enhance the representation of phenotypic relationships, embedding-based approaches have been proposed. For instance, [30] introduces an enriched embedding framework that integrates heterogeneous biomedical resources into the Human Phenotype Ontology (HPO), enabling improved detection of phenotypic similarity and supporting tasks such as patient stratification and phenotype relevance analysis.

Current rare disease databases are largely literature-based and may not fully support clinical diagnosis. Another study enhances such resources by extracting phenotype–disease relationships directly from EHRs using association rule mining [31]. Nevertheless, traditional phenotype KGs often lack detail, focusing mainly on core concepts while ignoring descriptive attributes. To overcome this, the PhenoSSU model introduces a fine-grained "entity-attribute-value" structure that captures richer semantic information about disease phenotypes [32]. A hybrid method combining concept recognition and ML was used to automate graph construction.

Biomedical systems like ROBOKOP further highlight the ability of KGs to generate mechanistic hypotheses by linking heterogeneous biomedical entities and uncovering hidden relationships [33]. Personal Health KGs (PHKGs) extend KG applications toward individualized healthcare by integrating multi-source personal health data, including EHRs, wearable devices and social determinants of health [34]. Further, despite their success in broader biomedical domains, KGs remain underexplored in psychiatry. A comprehensive review in [35] highlights their potential to support biomarker discovery, patient stratification and personalized interventions in mental health research. Recent developments, including MDKG [36] and psychosomatic disorder graphs [37], demonstrate how KGs can integrate fragmented evidence and uncover complex relationships between symptoms, treatments and biological mechanisms.

Additionally, predictive models combining KGs with ML techniques have shown improved performance in detecting depression-related behaviors and enhancing mental health monitoring systems [38]. Also, recent studies have demonstrated the effectiveness of KGs in integrating phenotypic, clinical and molecular data to uncover complex disease mechanisms. For instance, [39] constructs a biomedical KG that links patient symptoms with molecular pathways and gene interactions, enabling the identification of clinically relevant associations and supporting hypothesis generation for disease mechanisms.

Beyond clinical settings, KGs are increasingly applied in public health and interdisciplinary domains. The eKG framework [40] leverages LLMs to extract epidemiological insights from WHO reports, enabling structured disease surveillance and outbreak analysis. Similarly, KGs have been used in recommender systems to promote healthier behaviors, such as guiding users toward nutritious food choices through graph-based learning [41]. In engineering and industrial contexts, KG-based approaches support tasks such as equipment health monitoring and resilience assessment. For example, KGMLN enables few-shot learning for health status prediction by embedding structured knowledge into ML models [42], while domain-specific graphs support disaster resilience analysis in built environments [19].

While KGs effectively capture relational structures, they typically lack explicit modeling of causal dependencies and probabilistic uncertainty. Existing causal discovery, LLMs and probabilistic graphical models provide complementary capabilities [43], [44], [45]; however, their integration with phenotype-driven KG representations remains limited in current research [46], [47]. Despite these advances, most existing KG-based approaches remain primarily predictive or descriptive, focusing on identifying known relationships or improving classification and ranking performance. Limited attention has been given to the systematic identification of missing, weakly supported or unexplored relationships, particularly under uncertainty and heterogeneous population structures. Furthermore, current methods rarely integrate phenotype-driven abstraction with causal and probabilistic reasoning to support hypothesis generation. This gap motivates the need for a unified framework that combines KG enrichment with evidence-aware and phenotype-specific reasoning mechanisms.

## 3. Methodology

The proposed methodology introduces a phenotype-driven and evidence-governed framework for KG enrichment that shifts the focus from purely confirmatory analysis to structured hypothesis discovery. The framework integrates graph representation learning, hypothesis generation and scientific evidence extraction to iteratively refine a structured knowledge base. The core idea is to transform heterogeneous user data into interpretable phenotype representations, which are then used to generate hypotheses, retrieve relevant scientific literature and extract structured claims for controlled graph expansion. Formally, the system operates on a time-dependent user state $S_t$ represented as a heterogeneous graph and evolves a KG $G_t$ through a sequence of phenotype-driven enrichment steps. The process illustrated in Figure 1 begins by transforming data into graph-based representations and discovering latent phenotypes through GNN-based clustering. For each phenotype, causal relationships are inferred using NOTEARS and complemented with Bayesian Networks (BN) to capture probabilistic dependencies. These structured representations constrain LLMs to generate scientifically plausible hypotheses, which are then validated against both data-derived structures and evidence retrieved from scientific literature. A significant innovation is the introduction of a novelty–plausibility scoring (NPS) mechanism that prioritizes relationships that are structurally supported yet underexplored in the literature. The framework formulates graph expansion as a multi-objective optimization problem, selecting Pareto-optimal claims that balance relevance, validation and novelty, thereby enabling targeted and reliable KG enrichment.

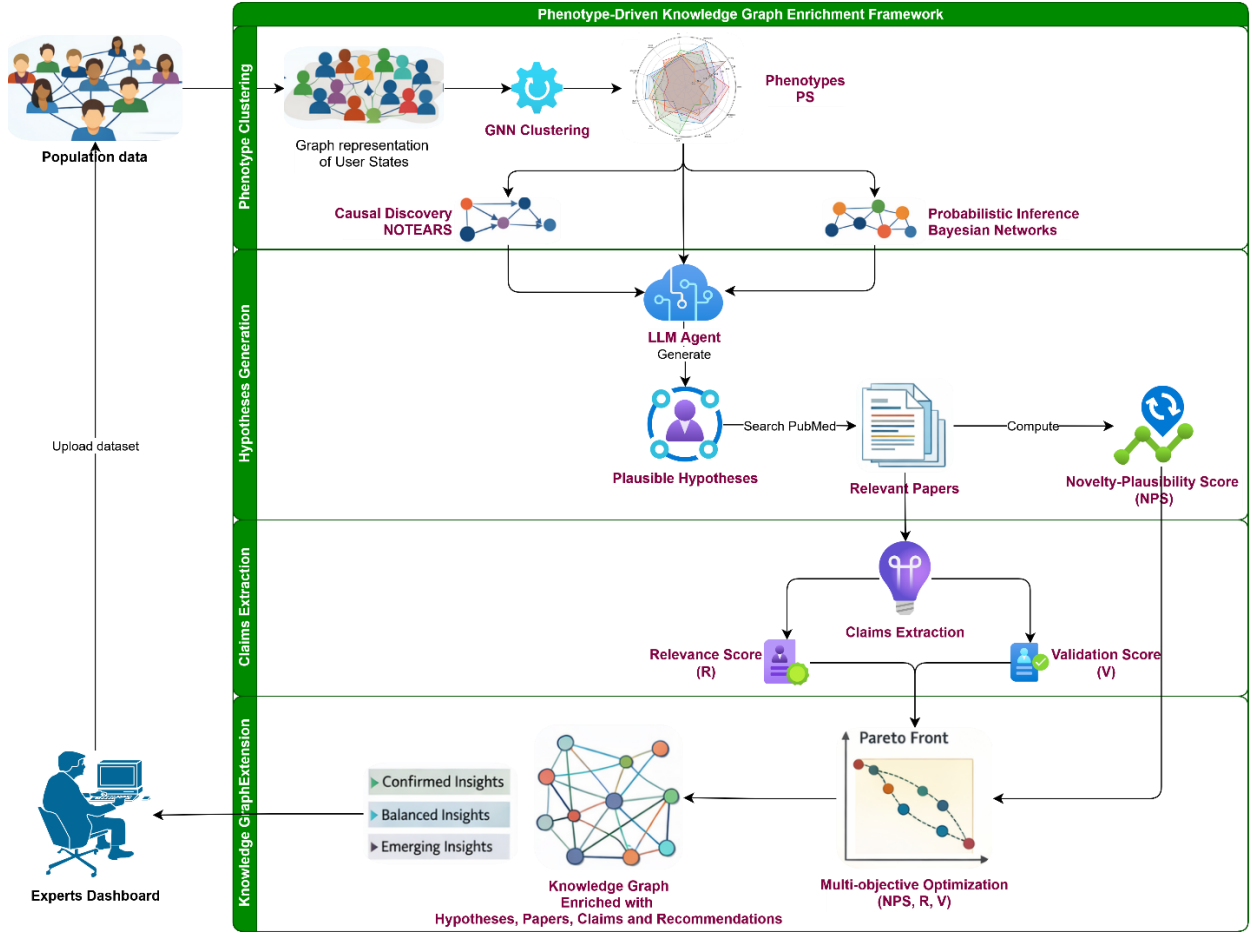

Figure 1. Methodology

### 3.1. Phenotyping pipeline using GNN and graph expansion
#### 3.1.1 State graph construction

At each time step $t$, the user state is represented as a heterogeneous graph $S_t = (X_t, E_t, V_t)$ where $X_t$ is the set of nodes (features $f$), $E_t \subseteq X_t \times X_t$ is the set of edges (relationships) and $V_t \in \mathbb{R}^{|X_t| \times f}$ is the node feature matrix. Features nodes may represent abstract entities such as symptoms, behaviors, medication, measurements, or contextual factors, while edges encode observed or inferred relationships. The user state is encoded using a GNN function which fits a latent representation for each node by aggregating both feature information and structural context. Formally, the node embeddings are obtained as:

$$Z_t = \text{GNN}(S_t) \in \mathbb{R}^{|X_t| \times h} \quad (1)$$

where $Z_t$ denotes the learned node embeddings and $h$ is the embedding dimension. To obtain a compact representation of the entire graph, a permutation-invariant readout function is applied over the node embeddings:

$$\mathcal{Z} = \text{READOUT}(Z_t) \quad (2)$$

where $\mathcal{Z} \in \mathbb{R}^h$ is the graph-level embedding summarizing the global structure and features of the user state.

#### 3.1.2 Phenotype discovery through GNN-based clustering

The purpose of this step is to identify recurrent cross-domain configurations of states that cannot be captured by isolated features alone. Because the graph embedding $\mathcal{Z}$ encodes structural, semantic and contextual information, clustering in this space discovers phenotypes that reflect both node attributes and inter-node dependencies. Thus, a clustering function $\Phi(\cdot)$ partitions the embedding space into $K$ phenotype groups $\Phi(Z) = \{P_1, P_2, \ldots, P_K\}$ where each phenotype $P_k$ is a subset of state embeddings:

$$P_k \subseteq Z, \bigcup_{k=1}^{K} P_k = Z \qquad (3)$$

and each state belongs either to one phenotype or to a soft membership distribution across phenotypes. In the hard-clustering case, the phenotype assignment of state $S_t$ is performed by minimizing the distances:

$$P_k(S_t) = \arg\min_k \; dist(Z_t, \mu_k) \qquad (4)$$

where $\mu_k$ is the centroid of phenotype $P_k$.

In the soft assignment case, the state is associated with a probability distribution over phenotypes:

$$\pi_t = (\pi_{t1}, \pi_{t2}, \ldots, \pi_{tK}), \sum_{k=1}^{K} \pi_{tk} = 1 \qquad (5)$$

where $\pi_{tk}$ denotes the degree to which state $S_t$ belongs to phenotype $P_k$. Therefore, a state is associated with multiple phenotypes, since state boundaries are often diffuse rather than discrete.

A set of standard phenotypes (SP) grounded in prior literature (e.g., anxiety, depression, stress, sleep disturbance, emotional eating and burnout) are used as interpretable abstractions of recurrent co-occurring patterns in the data. Let's define the set of standard phenotype nodes $SP = \{sp_1, sp_2, \ldots, sp_m\}$. Each standard phenotype $sp_m$ is characterized by a canonical node subset, edge pattern and feature distribution. Because discovered phenotypes (clusters) and standard phenotypes are not expected to align in a one-to-one manner, the fuzzy mapping is applied. For each discovered phenotype $P_k$, a similarity score $s_{km}$ is computed with respect to each standard phenotype $sp_m$. These scores are then normalized using softmax:

$$P(sp_m \mid P_k) = \frac{e^{s_{km}/T}}{\sum_{j=1}^{m} e^{s_{jm}/T}} \qquad (6)$$

where $T > 0$ is a temperature parameter controlling the sharpness of the distribution. The resulting vector $\Omega_k = (P(sp_1 \mid P_k), \ldots, P(sp_m \mid P_k))$ represents the standard phenotype composition of phenotype $P_k$. Therefore, each discovered phenotype is interpreted as a probabilistic mixture of known standard patterns rather than being forced into a single discrete label.

### 3.1.3. Determine relationships between the nodes of each phenotype (P)

Each phenotype $P_k$ is treated as a local subpopulation with its own internal structure. Instead of assuming that relationships between features are globally stable across the entire dataset, the model considers that dependencies may vary from one phenotype to another. Therefore, for each phenotype $P_k$, a specific dataset $X^{(k)}$ is obtained by collecting the feature representations of the states assigned to that phenotype. This dataset is used to estimate both directional dependencies and probabilistic relationships among the features. If phenotype $P_k$ contains $n_k$ states and each state is described by $f$ variables, then $X^{(k)} \in \mathbb{R}^{n_k \times f}$. Thus, the matrix becomes the input for both causal discovery and Bayesian modeling.

***Phenotype-specific causal discovery using NOTEARS***

To determine direct dependencies among the variables within each phenotype, NOTEARS algorithm is applied to the phenotype-specific matrix $X^{(k)}$. The aim is to infer a directed acyclic graph (DAG) that represents candidate causal relationships among the features for that phenotype. Let $W^{(k)} \in \mathbb{R}^{f \times f}$ be the weighted adjacency matrix for phenotype $P_k$, where $w_{(i-ii)}^{(k)}$ denotes the strength of the directed effect from features $X_{(i)}$ to $X_{(ii)}$. NOTEARS estimates $W^{(k)}$ by minimizing a reconstruction loss subject to an acyclicity constraint:

$$\min_{W^{(k)}} \frac{1}{2n_k} \| X^{(k)} - X^{(k)} W^{(k)} \|_F^2 + \lambda \| W^{(k)} \|_1 \qquad (7)$$

subject to:

$$h(W^{(k)}) = tr\left(e^{W^{(k)} \circ W^{(k)}}\right) - f = 0 \tag{8}$$

where:
- $tr(\cdot)$ denotes the trace of the square matrix, defined as the sum of its diagonal elements,
- $e^{(\cdot)}$ is the matrix exponential,
- $\|\cdot\|_F$ is the Frobenius norm,
- $\|\cdot\|_1$ promotes sparsity,
- $\circ$ denotes element-wise multiplication,
- $h(W^{(k)}) = 0$ ensures that the resulting graph is acyclic.

The output is a phenotype-specific causal graph $C^{(k)} = (X^{(k)}, E^{(k)})$, where $X^{(k)}$ is the set of features and $E^{(k)}$ is the set of directed edges inferred within phenotype $P_k$. These edges are interpreted as candidate causal dependencies and later used as structural evidence for claim validation.

*Bayesian networks for phenotype-level probabilistic reasoning*

While NOTEARS provides directional structure, it does not quantify uncertainty in a directly usable way. Therefore, for each phenotype $P_k$, we train a Bayesian network on the same phenotype-specific set $X^{(k)}$. The BN captures conditional dependencies among the variables and allows probabilistic inference for individual states belonging to that phenotype. Let $\mathcal{B}^{(k)}$ denote the BN associated with phenotype $P_k$. If the feature set in $X^{(k)}$ is $\{X_{(1)}, X_{(2)}, \ldots, X_{(f)}\}$, then the joint distribution factorizes according to the phenotype-specific graph structure as:

$$P(X_{(1)}, X_{(2)}, \ldots, X_{(f)} \mid P_k) = \prod_{i=1}^{f} P(X_{(i)} \mid \text{Pa}^{(k)}(X_{(i)})) \tag{9}$$

where $\text{Pa}^{(k)}(X_{(i)})$ denotes the set of parent variables of feature $(i)$ in phenotype $P_k$.

For a new state $S_t$ assigned to phenotype $P_k$, the BN is used to compute posterior probabilities of target variables conditioned on the observed evidence $P(Y^k \mid X^k, P_k)$, where $Y^k$ is a target variable of interest, such as a symptom, medication or risk state.

### 3.1.4. Generate structured hypotheses and claims extraction

For each phenotype $P_k$, we construct a phenotype state representation $PS_k = f_P(S_t, P_k)$, where $f_P$ is a phenotype state encoder that captures the core structure and dependencies of the phenotype-specific subgraph. Thus, the output $PS_k = \{\tilde{X}_k, \tilde{E}_k\}$ is a structured representation where $\tilde{X}_k$ are dominant features extracted from $X^{(k)}$ and $\tilde{E}_k$ are salient relationships including contextual constraints (e.g., temporal, environmental, or population-related).

A hypothesis generation function $\mathcal{L}_H$ is implemented using an LLM, which takes as input the phenotype state representation $PS_k$, causal graph $C^{(k)}$ and probabilistic BN representation $\mathcal{B}^{(k)}$. Using this combination, LLM is constraint to operate within a *causally coherent hypothesis space*. It prevents trivial or logically inconsistent hypotheses such as reversing causal direction (e.g., target → root) or proposing relationships that contradict the learned graph. The hypothesis function $\mathcal{L}_H$ produces a set of structured hypotheses:

$$\mathcal{H}_k = \mathcal{L}_H(PS_k, C^{(k)}, \mathcal{B}^{(k)}) \tag{10}$$

Thus, for each $PS_k$, a set of cross-domain hypotheses $\mathcal{H}_k = \{h_i \mid h_i = (Pop_i, I_i, O_i)\}$ is generated using PICO (*Population, Intervention, Comparison, Outcome*) like style that enables the transformation of research hypotheses into PubMed[1] queries. Thus, each hypothesis $h_i$ is a structured tuple with the following elements: population or context $Pop_i$, intervention or exposure $I_i$, outcome $O_i$.

For each hypothesis $h_i$, a query function retrieves a set of documents $D(h_i)$ from PubMed. Each document $d_p \in D(h_i)$ is assigned with a matching score:

$$M(d_p) = \alpha_1 \cdot f_{rel}(d_p, h_i) + \alpha_2 \cdot f_{rec}(d_p) \tag{11}$$

where $\alpha_1 + \alpha_2 = 1$ are positive weighting coefficients.

The scoring functions are defined as follows:
- $f_{rel}(d_i, h_i)$ indicates semantic relevance, computed as the similarity between the document content (e.g., title and abstract embeddings) and the hypothesis $h_i$ using cosine similarity in a shared embedding space.
- $f_{rec}(d_i)$ expresses recency and number of citations, defined as a time-based score that prioritizes more recent publications using an exponential decay function over publication year.

The retrieved documents are then sorted in descending order based on the matching score, and only those exceeding a predefined threshold $\tau_d$ are retained for further processing.

*Discovery of underexplored and latent relationships*

Beyond the identification of data-supported causal relationships, the proposed framework aims to discover underexplored, weakly supported, or potentially novel hypotheses. The objective is to identify relationships that are simultaneously plausible given the data-driven statistics and existing knowledge. Let $(X_{(i)}, X_{(ii)})$ denote a candidate relationship between features $X_{(i)}$ and $X_{(ii)}$ that lead to hypothesis $h_i(X_{(i)}, X_{(ii)})$. For each pair, a novelty-plausibility score $NPS(h_i)$ that combines the following five complementary components is proposed:

*1. Structural plausibility score $S_{struct}$* quantifies the strength of the relationship based on the causal graph $C^{(k)}$. It considers low confidence but structurally valid edges identified by the causal discovery process, corresponding to relationships that appear in the learned graph but are associated with relatively small edge weights or low statistical strength. While such edges may be disregarded in purely predictive settings, they represent potentially meaningful but weak signals, which may arise due to limited sample size, measurement noise, or context-specific effects. Retaining and analyzing these edges allows the framework to highlight relationships that are supported by the data structure but not yet strongly established, thereby serving as candidates for further investigation.

$$S_{struct}(h_i) = |w^{(k)}_{(i-ii)}| \tag{12}$$

The chains of dependencies of the form $X_{(i)} \rightarrow X_{(ii)} \rightarrow Y^k$ are also explored where no direct edge between feature $X_{(i)}$ and outcome $Y^k$ is present. Such configurations suggest the existence of mediated or latent mechanisms $X_{(ii)}$, where the influence of a variable propagates through intermediate factors. It considers the generation of hypotheses regarding hidden or unmodeled direct relationships, as well as the role of mediators in shaping observed outcomes. For these indirect relationships (multi-hop paths) $S_{path}$ includes the strongest path $\mathcal{P}_j$:

$$S_{path}(h_i) = \max_{\mathcal{P}_j} \prod_{(i,ii) \in \mathcal{P}_j} |w^{(k)}_{(i-ii)}| \tag{13}$$

---

[1] https://pubmed.ncbi.nlm.nih.gov/

*2. Probabilistic influence score $S_{prob}$* uses the BN outputs and measures how much conditioning on $X_{(i)}$ changes the distribution of $X_{(ii)}$. Let $P(X_{(ii)})$ and $P(X_{(ii)} | X_{(i)})$ denote the marginal and conditional distributions, respectively, the $S_{prob}$ is expressed as:

$$S_{prob}(h_i) = D_{KL}(P(X_{(ii)} | X_{(i)}) \parallel P(X_{(ii)})) \tag{14}$$

where $D_{KL}$ is the Kullback–Leibler divergence.

*3. Markov Blanket (MB) deviation score $S_{MB}$* identifies structurally unexpected relationships based on MB membership and highlights potentially missing or indirect dependencies. In $\mathcal{B}^{(k)}$, the MB of a node defines the minimal set of variables that render it conditionally independent from the rest of the network. Variables that are not included in this set are assumed to have no direct influence. However, when a variable exhibits strong associations elsewhere in the network but is absent from the MB of a target, this may indicate unobserved confounding, omitted edges, or context-dependent interactions. Such discrepancies are treated as signals of structural incompleteness, prompting the formulation of exploratory hypothesis $h_i (X_{(i)}, X_{(ii)})$:

$$S_{MB}(h_i) = \begin{cases} 0, & \text{if } X_{(i)} \in MB(X_{(ii)}) \\ 1, & \text{otherwise} \end{cases} \tag{15}$$

*4. Cross-phenotype variability score $S_{var}$* captures heterogeneity across phenotypes, indicating context-dependent relationships. Relationships that are present in one phenotype but absent or reversed in another suggest the presence of heterogeneous or conditional effects. These differences may reflect underlying moderation mechanisms, where the strength or direction of a relationship depends on the broader behavioral or physiological context.

$$S_{var}(h_i) = \text{Var}(w_{(i-ii)}^{(1)}, w_{(i-ii)}^{(2)}, \dots, w_{(i-ii)}^{(k)}) \tag{16}$$

*5. Literature novelty score $S_{lit}$* assigns higher scores to relationships that are rarely or weakly supported in the literature. Relationships that are strongly supported by the learned causal and probabilistic models but rarely mentioned in the extracted literature are considered candidates for novel or underreported findings. Conversely, relationships frequently reported in the literature but weakly supported in the data may indicate contextual differences, population-specific effects, or potential biases. The dual perspective enables the prioritization of hypotheses based on both empirical plausibility and novelty. Let $L_{(i-ii)}$ denote the normalized frequency or confidence of the relationship $(X_{(i)}, X_{(ii)})$ in the searched literature, then $S_{lit}$ is expressed as:

$$S_{lit}(h_i) = 1 - L_{(i-ii)} \tag{17}$$

The NPS score is defined as a weighted combination of the above components:

$$NPS(h_i) = \theta_1 S_{struct}(h_i) + \theta_2 S_{path}(h_i) + \theta_3 S_{prob}(h_i) + \theta_4 S_{MB}(h_i) + \theta_5 S_{var}(h_i) + \theta_6 S_{lit}(h_i) \tag{18}$$

where $\theta_1 \dots \theta_6$ are non-negative weights such that $\theta_1 + \theta_2 + \theta_3 + \theta_4 + \theta_5 + \theta_6 = 1$.

The NPS reads as a "structurally plausible yet underexplored" score as synthetized in Table 1.

Table 1. NPS components

| Component | What it captures | Meaning |
|---|---|---|
| $S_{struct}$ | Direct causal edge weight | Confirm *source→target* exists |
| $S_{path}$ | Indirect path strength | Even indirect evidence supports it |
| $S_{prob}$ | Bayesian conditional influence | Probabilistic dependency is real |
| $S_{MB}$ | Markov blanket deviation | Source is statistically relevant to target |
| $S_{var}$ | Cross-cluster variability | The effect isn't uniform, worth investigating per population |
| $S_{lit}$ | Inverse literature support | Literature hasn't covered this yet or poor literature coverage (low number of papers) |

The proposed NPS score balances two competing objectives: i) plausibility, captured by structural and probabilistic components and ii) novelty, captured by literature scarcity and structural gaps. High values of $NPS(h_i)$ indicate relationships that are supported by data (directly or indirectly), influential in probabilistic terms, variable across phenotypes and underrepresented in existing knowledge. Such relationships are prioritized as candidate hypotheses for further exploration, while the generated claims are candidates for the KG expansion.

*Claims extraction*

The claim extraction is performed for each hypothesis only on documents that are relevant based on the predefined threshold $\tau_d$. Each selected document $d_p$ is processed by the LLM ($\mathcal{L}_C$) to extract structured claims $C_d = \{c_j | c_j = (s_j, r_j, o_j, m_j)\}$, where each claim $c_j$ is represented by a subject entity $s_j$, a relation $r_j$ between entities, object entity $o_j$ and metadata $m_j$ containing evidence type, confidence, context and recommendations/interventions.

$$C_d \leftarrow \mathcal{L}_C(d_p, PS_k, \mathcal{C}^{(k)}, \mathcal{B}^{(k)}) \tag{19}$$

The claims are ranked using a relevance score $R(c_j)$ that considers the LLM confidence, population alignment with the PS, including the evidence strength of the study (observational, review, meta-analysis or randomized controlled trials - RCTs). Also, the extracted claims are not accepted purely based on textual extraction confidence of the LLM. Instead, they are validated against the phenotype-specific causal and probabilistic layers to ensure that claims are consistent not only with literature but also with the data-derived structure of the target phenotype. Thus, the relevance and validation scores of the claim $c_j$ are computed as:

$$R(c_j) = \omega_1 \cdot f_{LLM}(c_j, d_p) + \omega_2 \cdot f_{pop}(c_j, d_p) + \omega_3 \cdot f_{ev}(c_j, d_p) \tag{20}$$

$$\Upsilon(c_j) = \beta_1 \cdot f_{causal}(c_j, P_k) + \beta_2 \cdot f_{BN}(c_j, S_t, P_k) \tag{21}$$

where:
- $\omega_1, \omega_2, \omega_3 \geq 0$ and $\omega_1 + \omega_2 + \omega_3 = 1$ and $\beta_1, \beta_2 \geq 0$ and $\beta_1 + \beta_2 = 1$ are weighting coefficients
- $f_{LLM}(c_j, d_p)$ represents the LLM confidence score based on its own judgment of evidence strength for the extracted claims
- $f_{pop}(c_j, d_p)$ measures the population alignment or the degree of match between the population described in the document and the target population in the claim expressed by the phenotype state
- $f_{ev}(c_j, d_p)$ expresses the evidence strength from the methodological quality of the study, approximated using document metadata such as study type (e.g., randomized controlled trial, meta-analysis, observational study)
- $f_{causal}(c_j, P_k)$ denote the degree to which the subject-object relation is supported by $\mathcal{C}^{(k)}$
- $f_{BN}(c_j, S_t, P_k)$ denote the relevance of the claim in $\mathcal{B}^{(k)}$.

### 3.1.5 Multi-objective formulation of graph expansion

The graph expansion process is formulated as a multi-objective optimization problem, where each candidate claim represents a potential expansion decision. Rather than combining all criteria into a single scalar score, the framework considers multiple competing objectives that reflect the desired properties of KG enrichment. For each candidate claim $c_j$, three primary objectives are defined:
- Relevance $R(c_j)$ captures alignment with the hypothesis and phenotype context,
- Validation $\Upsilon(c_j)$ ensures consistency with the phenotype-specific causal and probabilistic structures,

- Novelty–plausibility $NPS(h_i)$ prioritize underexplored yet structurally supported relationships in the hypothesis that generated the claim.

These objectives are inherently conflicting (Table 2): R rewards claim that *is* well-supported by retrieved papers (confidence, matched nodes, evidence type); ϒ rewards claim that *aligns* with the graph structure, NPS rewards candidates supported by the graph, but the literature is *thin*.

Table 2. Conflicting objectives of the multi-objective problem

| Case | R | ϒ | NPS | Interpretation |
|---|---|---|---|---|
| 1 | High | High | High | New discovery - strong literature claim that also sits on an underexplored structural relationship (rare but most valuable) |
| 2 | Low | Low | High | Pure structural novelty - the graph sees something literature hasn't found yet. Worth generating new hypotheses about. |
| 3 | High | High | Low | Well-established knowledge - literature confirms what the graph shows, but it's already established |

Highly relevant and well-validated claims often correspond to well-established knowledge, resulting in low novelty. Conversely, highly novel claims may lack strong validation or may be weakly supported by existing evidence. Therefore, the expansion process cannot be reduced to a single optimal solution but instead requires identifying a set of Pareto-optimal candidates.

Let each candidate claim $c_j$ be represented as a vector in the objective space:
$$\mathcal{F}(c_j) = (R(c_j), \Upsilon(c_j), NPS(h_i | c_j \in h_i)) \qquad (22)$$

A claim $c_a$ is said to dominate another claim $c_b$ if:
$$f_s(c_a) \geq f_s(c_b) \forall s \in \{R, \Upsilon, NPS\} \text{ and } \exists\, s: f_s(c_a) > f_s(c_b) \qquad (23)$$

The set of Pareto-optimal claims consists of all non-dominated candidates:
$$C^* = \{c_j \in C \mid \nexists\, c_a \in C \text{ such that } c_a \succ c_j\} \qquad (24)$$

From the Pareto-optimal set $C^*$, a subset of claims is selected for KG expansion based on additional constraints such as minimum validation threshold $\tau_c$ and minimum matching threshold $\tau_d$ for documents:

The graph update is defined as:
$$G_{t+1} = G_t \cup \{c_j \in C^* \mid \Upsilon(c_j) \geq \tau_c\} \cup \{d_p \in D(h_i) \mid M(d_p) \geq \tau_d \wedge d_i \in D(h_i)\} \qquad (25)$$

where $D(h_i)$ denotes the set of documents supporting hypothesis $h_i$.

These claims represent optimal trade-offs between relevance, structural validity and novelty, considering that no selected claim can be improved in one objective without degrading another. Because the problem is defined over a finite set of candidate claims and involves only three objectives with limited constraints, the Pareto-optimal set can be computed using non-dominated sorting. Thus, no complex evolutionary optimization is required.

The main advantage of the Pareto-based formulation is that it filters out trivial or redundant knowledge. Claims that are highly relevant and well-supported but lack novelty such as well-known relationships (for e.g., *sugar increases glycemia* or *too much coffee may cause insomnia*) tend to be dominated by other candidates that achieve comparable relevance and validation while offering higher novelty. As a result, the Pareto front shifts the selection toward non-trivial, phenotype-specific and underexplored relationships, without explicitly requiring handcrafted thresholds for triviality. Since LLMs already encode common and frequently observed relationships with high confidence, prioritizing Pareto-optimal claims ensures that the graph focuses on regions of the knowledge space where grounding is most needed, thereby reducing hallucination risk and improving the quality of personalized recommendations.

The steps of the proposed model for KG expansion are expressed as pseudocode in Algorithm 1.

**Algorithm 1:** Phenotype-driven Knowledge Graph Expansion

| 1 | **Input:** Dataset of user states DS, standard phenotypes SP |
|---|---|
| 2 | **Output:** Expanded knowledge graph $G_{t+1}$ |
| 3 | **Parameters:** |
| 4 | $\omega_1, \omega_2, \omega_3 \geq 0 \mid \omega_1 + \omega_2 + \omega_3 = 1$; |
| 5 | $\beta_1, \beta_2 \geq 0 \mid \beta_1 + \beta_2 = 1$; |
| 6 | $\theta_1, \dots, \theta_6 \geq 0 \mid \theta_1 + \theta_2 + \theta_3 + \theta_4 + \theta_5 + \theta_6 = 1$ |

**Step 1: Data Processing**

| 7 | Encode categorial data: $DS' \leftarrow \text{LabelEncoder}(DS)$ |
|---|---|
| 8 | Transform $DS'$ into graph nodes $G_t \leftarrow \{S_t \mid S_t = (X_t, E_t, V_t)\}$ |
| 9 | Node embeddings: $Z_t = \text{GNN}(S_t) \in \mathbb{R}^{\lvert X_t \rvert \times h}$; $\mathcal{Z} = \text{READOUT}(Z_t)$ |

**Step 2: Phenotype Discovery**

| 10 | Select or estimate number of clusters $K$: $\{P_1, P_2, \dots, P_K\} \leftarrow \Phi(\mathcal{Z})$ |
|---|---|
| 11 | Associate $S_t$ to $P_k$ using soft assignment |
| 12 | Compute mapping to standard phenotypes: |

$$P(sp_m \mid P_k) = \frac{e^{s_{km}/T}}{\sum_{j=1}^{m} e^{s_{jm}/T}}$$

**Step 3: Determine relationships between the nodes of each phenotype (P)**

| 13 | **For** each phenotype $P_k$ **do** |
|---|---|
| | Select features $X^{(k)}$ |
| | Perform NOTEARS causal analysis $W^{(k)} \leftarrow \text{NOTEARS}(X^{(k)})$: |

$$\mathcal{C}^{(k)} = (X, E^{(k)})$$

Compute Bayesian network $\mathcal{B}^{(k)}$ based on $\mathcal{C}^{(k)}$:

$$\mathcal{B}^{(k)} \leftarrow BN(\mathcal{C}^{(k)})$$

**Step 4: Hypothesis Generation and Evaluation**

| 14 | **For** each phenotype $P_k$ **do** |
|---|---|
| | Construct state representation: $PS_k = f_P(S_t, P_k)$ |
| | Generate hypotheses and PIO query using LLM: |
| | $\mathcal{H}_k \leftarrow \mathcal{L}_H(PS_k, \mathcal{C}^{(k)}, \mathcal{B}^{(k)})$ where $\mathcal{H}_k = \{h_i \mid h_i = (Pop_i, I_i, O_i)\}$ |
| | For each hypothesis $h_i$: |
| | Retrieve documents $D(h_i)$ from PubMED |
| | Compute hypothesis scores (s): |

$$S_{struct}(h_i), S_{prob}(h_i), S_{MB}(h_i), S_{var}(h_i), S_{lit}(h_i)$$

Compute NPS score:

$$NPS(h_i) = \sum_s \theta_s S_s(h_i)$$

**For** each document $d_p \in D(h_i)$ **do**
  Compute matching score: $M(d_p) = \alpha_1 \cdot f_{rel}(d_p, h_i) + \alpha_2 \cdot f_{rec}(d_p)$
  Select $d_p$ based on the threshold $\tau_d$: $d_p \leftarrow d_p \mid M(d_p) \geq \tau_d$
  Extract claims from document $d_p$: $C_d \leftarrow \mathcal{L}_C(d_p, h_i) \mid C_d = \{c_j \mid c_j = (s_j, r_j, o_j, m_j)\}$
  **For** each claim $c_j \in C_d$ **do**
    Compute relevance and validation scores:

$$R(c_j) = \omega_1 \cdot f_{LLM}(c_j) + \omega_2 \cdot f_{pop}(c_j, d_i) + \omega_3 \cdot f_{ev}(c_j, d_i)$$
$$Y(c_j) = \beta_1 \cdot f_{causal}(c_j, P_k) + \beta_2 \cdot f_{BN}(c_j, S_t, P_k)$$

**Step 5: Multi-objective Graph Expansion**

| 15 | Represent each claim: |
|---|---|
| 16 | $\mathcal{F}(c_j) = (R(c_j), Y(c_j), NPS(h_i \mid c_j \in h_i))$ |
| 17 | Compute Pareto-front set: $C^* = \{c_j \in C \mid \nexists c_a \in C \text{ such that } c_a \succ c_j\}$ |
| 18 | Expand graph based on the selected claims: |
| 19 | $G_{t+1} = G_t \cup \{c_j \in C^* \mid Y(c_j) \geq \tau_c\} \cup \{d_p \in D(h_i) \mid M(d_p) \geq \tau_d \wedge d_i \in D(h_i)\}$ |

### 3.2. Online processing of new user states

For a newly observed user state $S'_t$, the system performs a matching procedure to determine whether the state can be associated with existing phenotypes or represents a novel pattern. Each new state is represented by a structured feature set and its corresponding latent representations $S'_t = \{X'_t, z'_t, \pi'_t\}$ where $X'_t$ denotes the raw feature vector, $z'_t$ is the GNN-derived embedding, and $\pi'_t$ is the fuzzy distribution over standard phenotypes. The similarity between the new state $S'_t$ and each phenotype $P_k$ is computed as a weighted combination of structural similarity in the embedding space and semantic alignment in the phenotype space:

$$\text{score}(S'_t, P_k) = \alpha \cdot \cos(z'_t, \mu_k) + (1-\alpha) \cdot \cos(\pi'_t, PS_k) \tag{26}$$

where:
- $\mu_k$ is the centroid embedding of phenotype $P_k$
- $\alpha \in [0,1]$ controls the relative importance of structural versus semantic similarity.

The assignment of $S'_t$ to existing phenotypes is determined based on the maximum similarity score:

$$score^* = \max_k score_k \tag{27}$$

The following thresholds are defined: a matching threshold $\tau_{\text{match}}$ and an anomaly threshold $\tau_{\text{anom}}$, with $\tau_{\text{anom}} < \tau_{\text{match}}$. If $score^* \geq \tau_{\text{match}}$ then the state is assigned to the corresponding phenotype $P_k$, else if $score^* < \tau_{\text{anom}}$ then the state is considered an anomaly and a new phenotype candidate is created ($P'_k$). If the score is between the two thresholds, then the state is treated as a soft match and associated with multiple phenotypes via a weighted assignment. In this case, a subset $\tilde{P}_k$ of phenotypes are selected based on $score_k$. The decision rule for new state assignment is expressed as:

$$\mathcal{A}(S'_t) = \begin{cases} \text{match}(P_{k^*}) & \text{if } score^* \geq \tau_{\text{match}} \\ \text{anomaly } P'_k & \text{if } score^* < \tau_{\text{anom}} \\ \text{soft\_match}(\tilde{P}_k) & \text{otherwise} \end{cases} \tag{28}$$

where $k^* = \arg\max_k score(S'_t, P_k)$,

To complement similarity-based matching, anomaly detection is performed in the embedding space using an unsupervised model such as Isolation Forest. Let $f_{\text{anom}}(z'_t) \in \{-1, 1\}$ denote the anomaly indicator, where $-1$ corresponds to anomalous states. A state is classified as anomalous if:

$$f_{\text{anom}}(z'_t) = -1 \wedge score^* < \tau_{\text{anom}} \tag{29}$$

The combined criterion considers that anomalies are detected both in terms of global distribution (embedding space) and lack of alignment with existing phenotypes.

A candidate representation of the new pattern is constructed as $P'_k = \{X'_t, z'_t, \pi'_t, n_c = 1\}$, where $n_c$ denotes the support count. For this candidate, an accelerated knowledge acquisition pipeline is triggered and a limited set of hypotheses are generated. The retrieval and claim extraction process for the candidates is constrained for reliability and computational efficiency and only a small number of hypotheses and documents are considered. All resulting knowledge is explicitly marked as exploratory, reflecting the uncertainty associated with the candidate phenotype. To prevent noise-driven expansion of the phenotype space, candidate phenotypes are maintained in a buffer. If a sufficient number of similar anomalous states accumulate ($n_c \geq \tau_{nc}$) then the candidate phenotype is promoted to a stable phenotype $P_{k+1} \leftarrow P'_k$.

## 4. Simulations
### 4.1 Dataset description

The proposed framework is evaluated on 2 datasets that reflect cross-domain interactions and multi-dimensional representation of the study population.

The first dataset (**DS1**) is the *Student Mental Health and Intervention Dataset*[2], which consists of 1,000 student records, each representing an individual snapshot of a student's state, described by 16 features spanning multiple domains. It integrates self-reported emotional indicators, physiological measurements, academic engagement metrics and behavioral attributes, alongside textual emotional reflections. Emotional wellbeing is captured through variables such as *stress*, *anxiety* and *depression levels*, while physiological regulation is reflected by *heart rate variability* and *sleep duration*. Academic behavior is represented by *attendance rate*, *study hours*, *learning platform engagement level (LMS)* and *academic performance index*, while *social interaction scores* capture peer engagement. The dataset includes an *emotional journal* field, providing unstructured textual data that reflects students' subjective experiences and can be leveraged for semantic enrichment. The dataset contains two key target variables: *Mental_Health_Risk*, which classifies students into low, medium, or high-risk categories and *Personalized_Intervention_Strategy* related to tailored psychological support actions.

---
[2] https://www.kaggle.com/datasets/zara2099/student-mental-health-and-intervention-dataset

The second dataset (**DS2**) contains data from the *Study of Women's Health Across the Nation (SWAN) Visit 07 dataset (ICPSR 31901)*[3], a midlife women's health cohort comprising 2,413 participants assessed between 2003 and 2005, spanning menopausal, psychosocial, cognitive, behavioral and biological domains. Starting from the original ICPSR tabular data with more than 1000 columns, we standardized field names from the codebook and transformed the high-dimensional visit-level data into a menopause-state dataset in which each participant is represented as a compact state vector of 39 features. These features capture i) menopause cross-domain interactions, menopausal stage and hormone therapy exposure; ii) vasomotor, sleep, somatic, urogenital, depressive, anxiety and perceived stress burdens; iii) quality-of-life impairment; iv) life-event stress, discrimination, financial strain and healthcare access barriers; v) subjective and objective cognition; vi) adiposity, blood pressure, glycemic, lipid, inflammatory, bone and cardiometabolic risk; vii) chronic condition burden; viii) behavioral context such as smoking, alcohol intake, healthy behaviors and complementary therapy engagement.

For the following sections we detailed the DS1 results, while DS2 is included in the Supplementary material.

### 4.2 Phenotyping using GNN-based graph clustering

We evaluated the effectiveness of graph-based phenotyping using a GNN combined with spectral clustering and compared it against a hierarchical clustering baseline. The hierarchical method identified for DS1 three clusters of unequal size (n=454, 273, 273), while the GNN-based approach produced 6 clusters (n ranging from 127 to 199). Quantitative evaluation (Table 3) shows that the GNN-based approach outperforms hierarchical clustering in terms of cluster compactness and balance. The silhouette score improves from 0.0454 (hierarchical) to 0.0794 (GNN), indicating better separation in the latent space. Moreover, the standard deviation of cluster sizes is significantly reduced (85.3 vs. 21.5), demonstrating that GNN clustering avoids the formation of dominant clusters and captures more evenly distributed subpopulations.

Table 3. Clustering Performance Comparison

| Metric | Hierarchical | GNN + Spectral |
| --- | --- | --- |
| Number of clusters | 3 | 6 |
| Silhouette score | 0.0454 | 0.0794 |
| Cluster size std | 85.3 | 21.5 |
| Min cluster size | 273 | 127 |
| Max cluster size | 454 | 199 |
| Mean SP entropy (bits) | 2.923 | 2.469 |

Hierarchical clustering produces coarse partitions primarily driven by dominant psychological variables such as stress, anxiety and depression. As shown in Figure 2, these clusters exhibit substantial overlaps across dimensions, limiting interpretability. In contrast, GNN-based clustering shows multi-dimensional profiles with distinct combination of psychological, behavioral and contextual features.

---

[3] https://www.swanstudy.org/

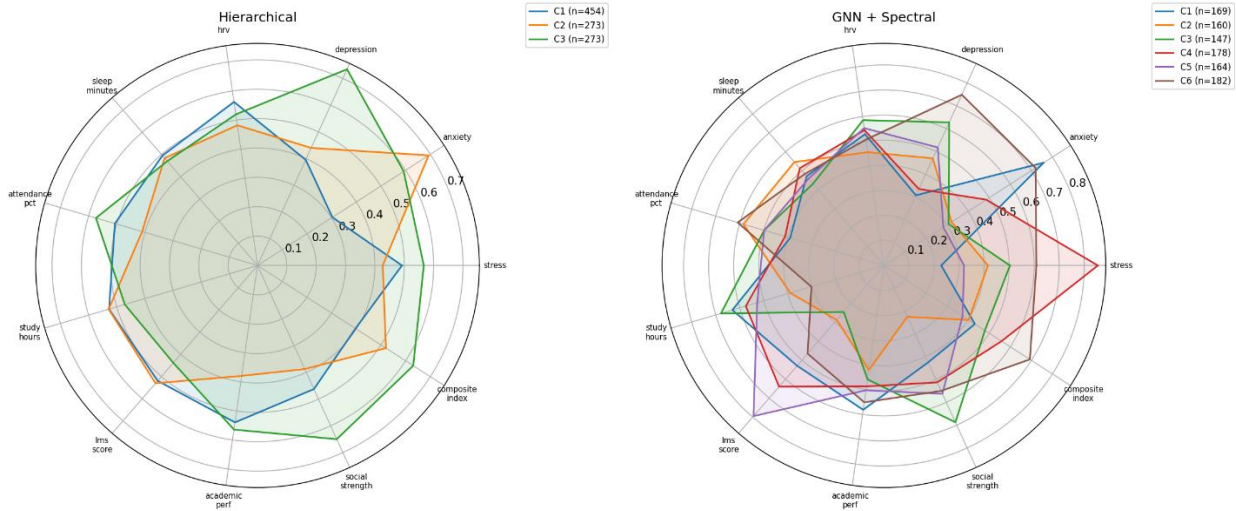

Figure 2. Radar plots - hierarchical vs GNN comparison

Cross-tabulation analysis (Figure 3) further confirms this divergence. Each hierarchical cluster is distributed across multiple GNN clusters, indicating that hierarchical clustering aggregates several finer-grained behavioral patterns into broader categories, whereas GNN decomposes them into more specific phenotypes.

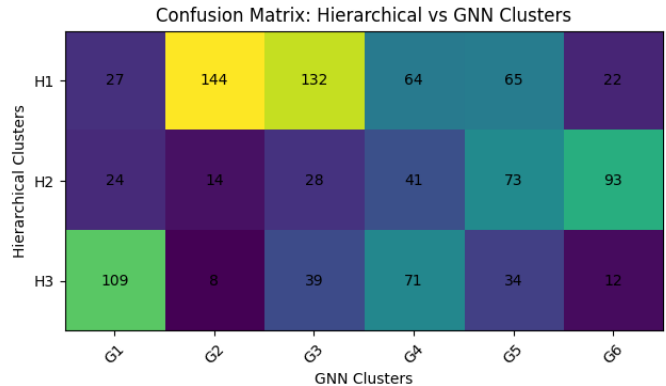

Figure 3. Cross-tabulation Hierarchical vs GNN clusters

To evaluate interpretability, clusters are mapped to predefined standard phenotypes (SP1–SP8 provided in the Annex). Fuzzy mapping to these SPs depicted in Figure 4 reveals that hierarchical clusters have weak and diffuse alignment, with low contrast between dominant and non-dominant phenotypes. In contrast, GNN clusters exhibit more selective phenotype signatures, including both positive and negative alignment values.

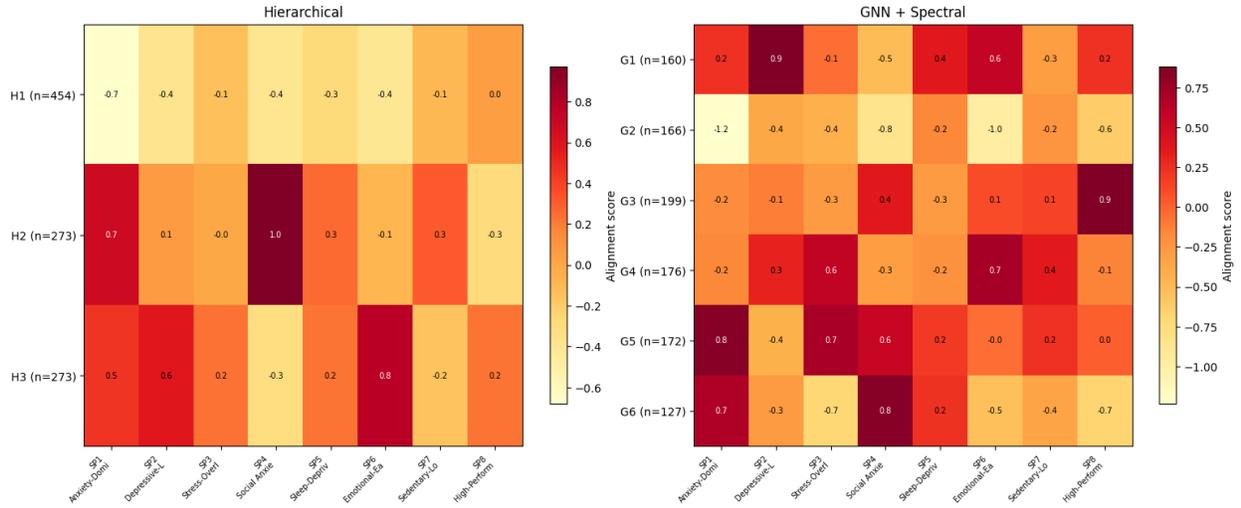

Figure 4. SP alignment heatmap

The results demonstrate that hierarchical clustering captures only coarse, low-dimensional separations in the data, whereas GNN-based clustering generates more interpretable behavioral phenotypes. Therefore, the GNN clusters provide a more realistic representation of human behavioral states, enabling downstream tasks such as causal inference, hypothesis generation and KG enrichment.

### 4.3 Causal and probabilistic analysis
#### 4.3.1 Phenotype-specific causal graphs

The phenotype-specific causal analysis shows high heterogeneity in the relationships between demographic, behavioral and outcome variables across phenotypes. For dataset DS1, Table A1 (Supplementary material) indicates that psychological variables (anxiety, stress and depression) act as central mediators linking exogenous factors (age and gender) to academic and behavioral outcomes. Across multiple phenotypes (C1, C2, C6), anxiety has a strong negative effect on academic performance, with the largest effect observed in C1. Similarly, stress plays a dominant role in C3, where it emerges as the primary determinant of academic performance. Thus, while the specific dominant psychological factor varies by phenotype, the same mechanism (psychological burden impairing performance) is stable across subpopulations.

Social and emotional variables form a tightly interconnected subsystem. Emotional journaling is associated with reduced anxiety (C1) and improved academic performance (C4), while also positively influencing social strength (C5). However, social strength itself exhibits complex and sometimes counterintuitive relationships, being positively associated with both depression (C1) and stress (C6). Thus, social engagement may capture both supportive interactions and socially induced pressures, depending on the phenotype. Therefore, while the high-level causal structure *exogenous → behavioral/psychological → outcomes* is consistent, the specific pathways and dominant mechanisms vary significantly across phenotypes.

#### 4.3.2 Bayesian inference at state level

To complement the causal structures, BN are applied to derive probabilistic relationships within each phenotype. The BN analysis provides a compact representation of conditional dependencies, where each variable is influenced only by its direct parents. Beyond local dependencies, the Markov blanket of a node enables the identification of minimal sufficient predictor sets for each target variable, improving both interpretability and practical applicability. Table A2 summarizes the key structural relationships for selected outcome variables in P1, including their direct causes, downstream effects and MB. Psychological variables (anxiety, stress, depression) do not act as terminal outcomes but rather as intermediate mediators, reinforcing their central role in shaping performance and engagement. Academic performance is primarily associated with anxiety, with its MB further including sleep duration and study behavior. Stress is influenced by social/behavioral factors, while depression is linked to social strength and emotional expression, forming a distinct subsystem that captures emotional and social dynamics.

### 4.3.3 Comparative analysis across phenotypes

To better understand the structural differences across identified phenotypes, we conduct a comparative analysis based on graph-level properties derived from both NOTEARS and BN representations. Thus, we examine the number of causal edges, sparsity levels and the complexity of probabilistic dependencies across clusters. Table 4 summarizes the structural characteristics of each phenotype, including the number of causal edges, sparsity and the corresponding BN complexity.

Table 4. Structural comparison of phenotypes

| Cluster | NOTEARS Edges | NOTEARS Sparsity | BN Nodes | BN Edges |
|---|---|---|---|---|
| 1 | 77 | 0.342 | 15 | 28 |
| 2 | 89 | 0.396 | 15 | 30 |
| 3 | 86 | 0.382 | 15 | 27 |
| 4 | 78 | 0.347 | 15 | 27 |
| 5 | 84 | 0.373 | 15 | 30 |
| 6 | 82 | 0.364 | 15 | 27 |

Phenotypes with a higher number of causal edges, such as C2 and C3, have more interconnected causal structures, suggesting that symptoms and behaviors are tightly coupled. This indicates that interventions in such groups may require holistic, system-level strategies, as changes in one variable are likely to propagate across the network. In contrast, phenotypes with fewer edges and lower sparsity, such as C1, display relatively simpler and more modular structures.

### 4.4 Hypothesis generation and claims extraction (LLM)

*Gemini 3 Flash* with the temperature set to 1 is used to generate hypotheses and evidence-based claims. In Figure A4 several hypotheses and PICO queries are provided for dataset DS1. The extracted claims are aligned with cluster-specific variables (PS) to ensure consistency with the identified phenotypes. The distribution of claims in Figure 5 indicates an alignment between confidence and relevance, with most points concentrated in the mid-to-high range of both dimensions (confidence ≈ 0.6–0.9, relevance ≈ 0.3–0.65). This suggests that most extracted claims are not only supported by the LLM (high confidence) but also meaningfully related to the underlying hypotheses and PS (moderate-to-high relevance). Larger bubbles that indicate higher validation are predominantly located in this central area, confirming that the framework effectively filters and retains claims that are both structurally and probabilistically consistent. Differences across clusters are visible but not extreme, implying stable performance of the method across phenotypes. A smaller number of claims appear with high confidence but lower relevance, highlighting potential generic or less context-specific findings, while very few low-confidence points are present, indicating good overall robustness of the extraction and validation pipeline.

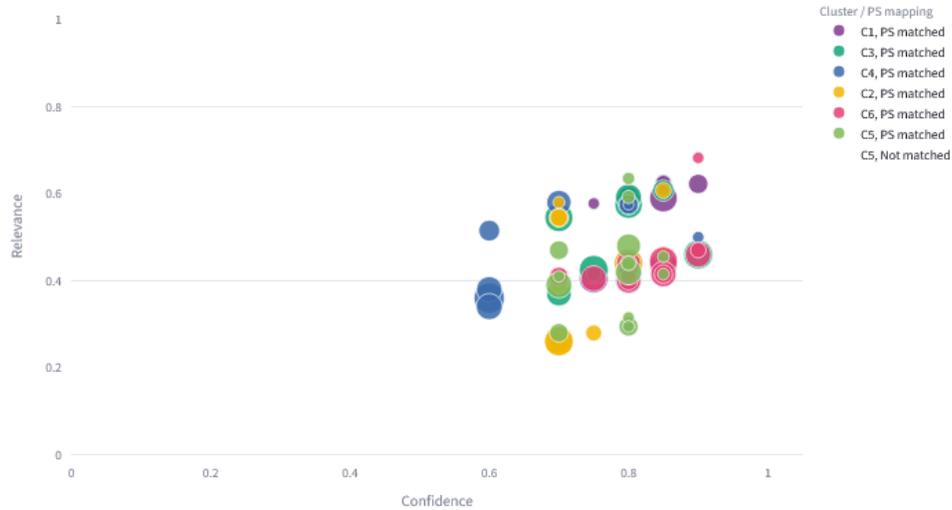

Figure 5. Claims' confidence and relevance scores distributed across clusters in DS1

The distribution of NPS components across clusters illustrated in Figure 6 reveals structural and contextual patterns that underpin the hypothesis evaluation process. Structural scores (*struct* and *path*) have moderate to high variability across clusters, indicating that both direct and indirect graph relationships contribute differently depending on the phenotype. The *prob* score remains generally low and concentrated, suggesting that high probabilistic dependencies are more selective and less uniformly present across clusters. In contrast, the *mb* score has high value across all clusters, highlighting its role in capturing statistically relevant dependencies within the graph. The *var* score shows a high spread, reflecting meaningful cross-cluster heterogeneity and reinforcing the importance of population-specific effects. The *lit* component is often skewed toward higher values, indicating that many candidate relationships are relatively underexplored in literature.

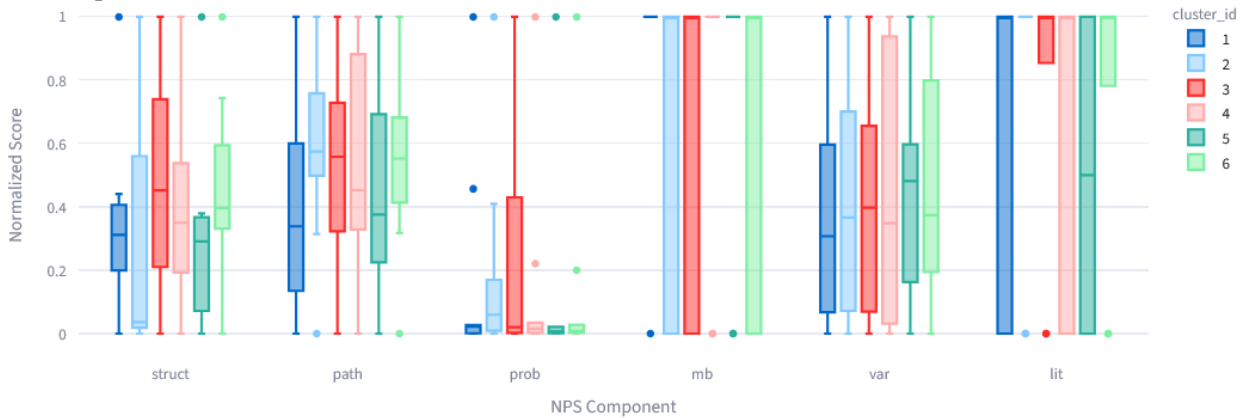

Figure 6. NPS distribution by cluster

The Pareto front in Figure 7 illustrates a subset of claims that achieve an optimal balance between relevance (R), validation (V) and novelty (NPS), distinguishing them from the broader set of non-optimal candidates.

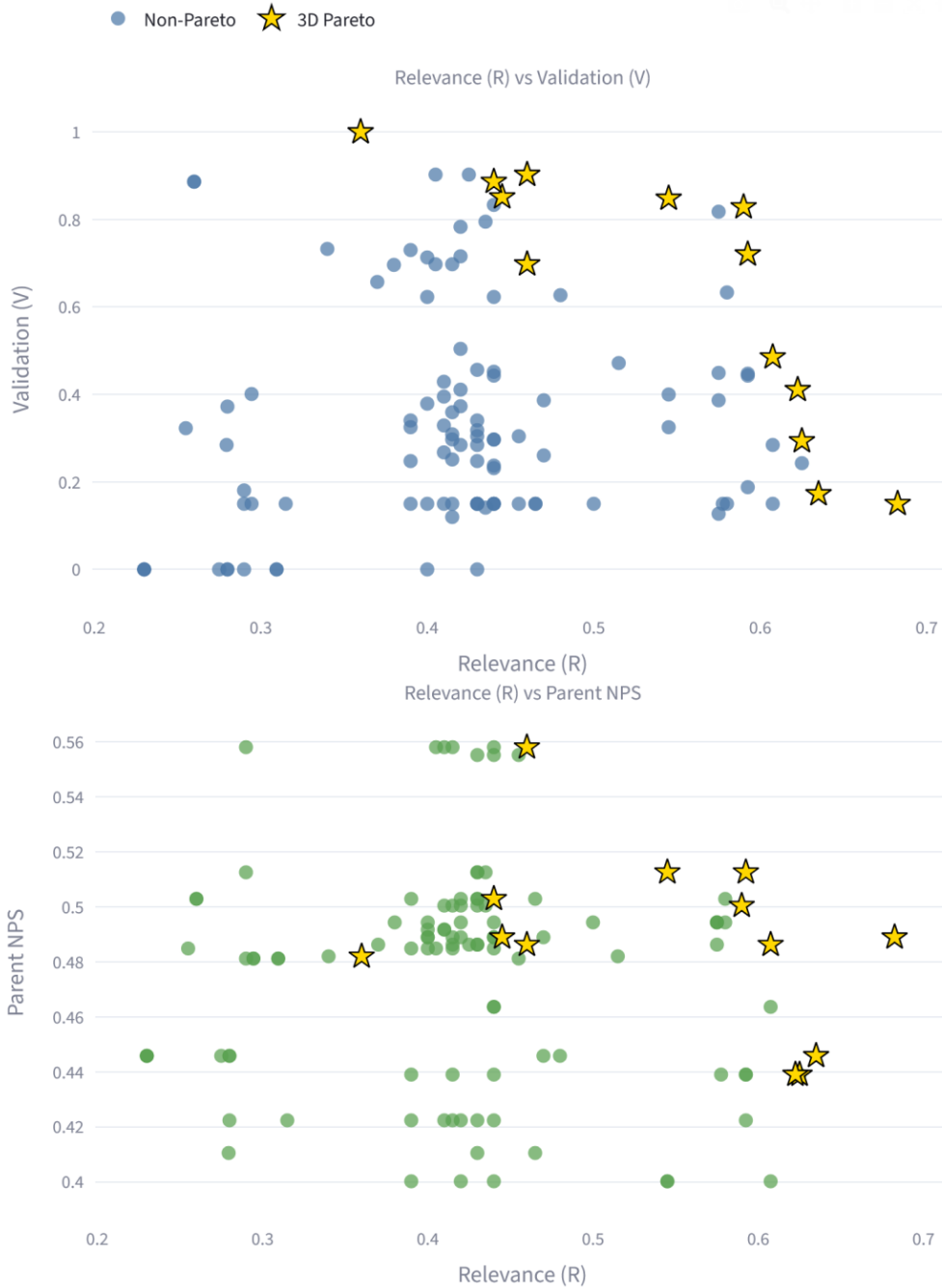

Figure 7. Pareto front representation of the claims

In the relevance–validation space, Pareto-optimal claims are predominantly located in the upper-right region, confirming that the selected claims simultaneously maximize contextual relevance and structural validation. Some Pareto points also extend into regions of moderate relevance but high validation, showing that strong structural support can compensate for slightly lower contextual alignment. In the relevance–NPS space, Pareto-optimal claims are distributed around mid-to-high NPS values, suggesting that the framework does not simply prioritize novelty, but rather balances it with evidence and validation. This confirms the effectiveness of the multi-objective formulation: instead of favoring only highly supported or only novel claims, the Pareto front captures diverse high-quality candidates, including well-supported findings and structurally plausible yet underexplored hypotheses.

The Pareto-optimal claims are evaluated by human experts (4 professors and 7 PhD students in the university) and their evaluation identifies three distinct and complementary categories of claims: i) a set of high-relevance & high-validation claims captures well-established relationships that are strongly supported by the literature, such as the association between female gender and higher anxiety levels or the role of parent and peer attachment in improving academic performance. These results demonstrate that the framework preserves evidence-backed knowledge; ii) a set of balanced claims having a trade-off across relevance, validation and novelty, representing structurally important relationships such as the link between stress and anxiety or the predictive role of attendance in academic achievement. This set forms the backbone of the KG, providing stability and generalizability across domains; iii) a set of high-novelty or emerging insights, characterized by relatively higher NPS and/or lower validation, including relationships such as the impact of robotic tutors on academic performance or the integration of AI in enhancing student engagement. These claims are structurally plausible yet less explored in the literature, illustrating the framework's ability to surface hypothesis-generating candidates. The KG expansion for DS1 is illustrated in Figure A6.

### 4.5 New users and anomaly detection

To evaluate the practical applicability of the proposed framework, we extend the analysis to unseen data by performing phenotype assignment and anomaly detection. For each new user state, we compute similarity scores with respect to the learned phenotype representations using a weighted combination of embedding similarity and structural proximity. Unlike hard clustering, this approach enables soft assignment, where a user state can exhibit partial similarity to multiple phenotypes. The results performed on 100 new states per dataset indicate that certain users demonstrate ambiguous membership, with comparable similarity scores across multiple clusters. For example, in Figure 8 the new user state shows proximity to both C3 (0.57) and C6 (0.35), suggesting overlapping behavioral characteristics between the two phenotypes.

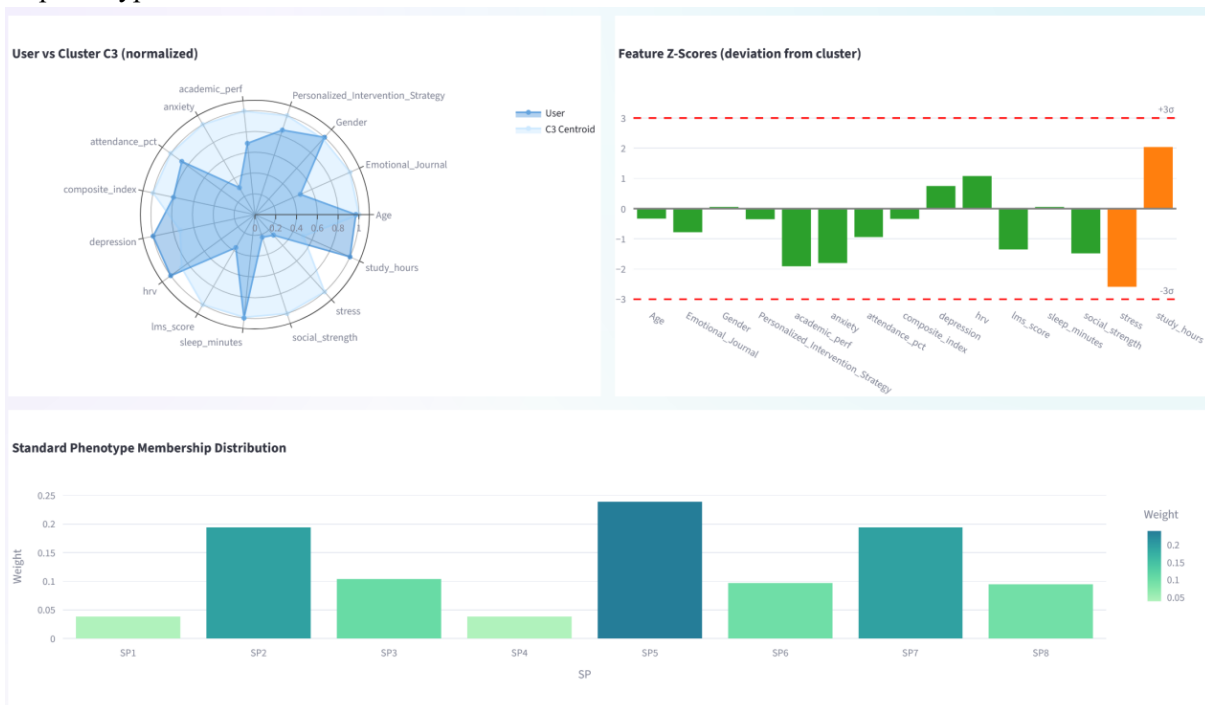

Figure 8. Soft match of a new user state

Users with low similarity scores and high anomaly values are flagged as out-of-distribution cases and are labeled as exploration patterns, representing potential new behavioral phenotypes not captured by existing phenotypes. An exploration pipeline is triggered to generate new hypotheses and extract claims from literature, and the knowledge graph is expanded accordingly (Figure A7).

**4.6 Ablation studies**

The ablation study synthetized in Table 5 was performed for both datasets and evaluates the impact of each component of the proposed framework in terms of (i) claim quality, (ii) structural consistency, (iii) graph expansion efficiency, and (iv) downstream utility for RAG systems.

Table 5. Ablation study comparing claim generation and graph expansion strategies

| Method | Plausibility score | | Novelty score | | Validation Score | | Relevance Score | | NPS | | RAG Recall@5 | | Hallucination Rate | |
|---|---|---|---|---|---|---|---|---|---|---|---|---|---|---|
| | DS1 | DS2 | DS1 | DS2 | DS1 | DS2 | DS1 | DS2 | DS1 | DS2 | DS1 | DS2 | DS1 | DS2 |
| **Rule-based** | 0.95 | 0.89 | 0.10 | 0.12 | 0.84 | 0.82 | 0.71 | 0.68 | 0.28 | 0.23 | 0.71 | 0.67 | 0.20 | 0.26 |
| **LLM only** | 0.37 | 0.32 | 0.56 | 0.61 | 0.28 | 0.34 | 0.31 | 0.27 | 0.34 | 0.29 | 0.83 | 0.78 | 0.27 | 0.32 |
| **Proposed** | 0.54 | 0.65 | 0.69 | 0.72 | 0.45 | 0.57 | 0.48 | 0.51 | 0.52 | 0.55 | 0.98 | 0.97 | 0.05 | 0.06 |

The rule-based method achieves very high plausibility (0.9), validation (0.8) and relevance (0.7), but exhibits extremely low novelty (0.1), resulting in a low overall NPS ($\cong 0.25$). The rule-based strategies are effective at recovering well-established and highly supported relationships but lack the ability to identify new or underexplored knowledge. In contrast, the LLM-only approach demonstrates higher novelty (>0.56) with low plausibility (<0.37), validation ($\cong 0.3$) and relevance ($\cong 0.3$), reflecting its tendency to generate more speculative or weakly grounded claims. The proposed method achieves a balanced performance across all dimensions, with moderate plausibility ($\cong 0.6$), improved novelty ($\cong 0.7$) and higher validation (>0.45) and relevance (0.5) compared to the LLM-only baseline, resulting in the highest NPS ($\cong 0.54$).

Therefore, the proposed framework enables the identification of claims that are both data-consistent and novel, avoiding the limitations of purely confirmatory or purely generative approaches. Related to RAG performance, the rule-based approach achieves moderate Recall@5 (0.7) with a relatively moderate hallucination rate ($\cong 0.23$), reflecting its reliance on well-established and constrained knowledge sources. The LLM-only method improves retrieval performance (Recall@5 $\cong$ 0.8) but exhibits the highest hallucination rate (>0.27), indicating that increased generative flexibility comes at the cost of reduced factual reliability. In contrast, the proposed method achieves the highest retrieval performance (Recall@5 $\cong 0.98$) while simultaneously maintaining the lowest hallucination rate (<0.06). Therefore, the integration of structural constraints, probabilistic validation and evidence-grounded retrieval not only enhances coverage but also significantly improves the factual consistency of generated outputs in retrieval-augmented settings.

**5. Conclusions**

The paper proposed a phenotype-driven and evidence-governed framework for KG extension, designed to move beyond traditional confirmatory analytics toward new hypothesis discovery. The framework integrates GNNs, causal discovery, Bayesian inference and LLM-based reasoning into a unified pipeline to enable identification of context-dependent, structurally grounded and underexplored relationships in heterogeneous population data.

The experimental results support for the research questions. For *RQ1*, the GNN-based phenotyping demonstrated the ability to capture cross-domain interactions, producing more balanced and interpretable clusters compared to hierarchical approaches. Regarding *RQ2*, the phenotype-specific causal and probabilistic analysis show heterogeneity across subpopulations, confirming that localized modeling improves the identification of context-dependent relationships. For *RQ3*, the integration of causal and BN graphs guided LLM hypothesis generation, resulting in claims that are both scientifically plausible and non-trivial, as reflected by the high concentration of claims with strong confidence, relevance and validation.

For *RQ4*, the proposed NPS and multi-objective optimization framework identify underexplored yet structurally supported relationships. The Pareto front analysis demonstrated that the method successfully balances relevance, validation and novelty, enabling the selection of diverse high-quality claims, including well-established knowledge, balanced insights and emerging hypotheses. This is further confirmed by expert evaluation, which validated the framework's ability to preserve known relationships while uncovering novel, hypothesis-generating patterns. Additionally, the ablation study shows that the proposed method achieves the best trade-off across plausibility, novelty, validation and relevance, leading to the highest overall NPS. In retrieval-augmented settings, it significantly improves performance (Recall@5=0.98) while reducing hallucination rates (0.05). For *RQ5*, the framework demonstrates adaptability to new user states through soft phenotype matching and anomaly detection and identify previously unseen behavioral patterns and supports incremental knowledge discovery.

Future work will focus on extending the framework to larger-scale real-world datasets, incorporating temporal dynamics in phenotype evolution and refining the human-in-the-loop feedback process to further enhance the interpretability and trustworthiness of discovered knowledge.

**Declarations**
**Funding**-This work was supported by a grant of the Ministry of Research, Innovation and Digitization, CNCS/CCCDI - UEFISCDI, project number COFUND-DUT-OPEN4CEC-1, within PNCDI IV. This project has been funded by UEFISCDI under the Driving Urban Transitions Partnership, which has been co-funded by the European Commission.
**Acknowledgement**-This work was supported by a grant of the Ministry of Research, Innovation and Digitization, CNCS/CCCDI - UEFISCDI, project number COFUND-DUT-OPEN4CEC-1, within PNCDI IV. This project has been funded by UEFISCDI under the Driving Urban Transitions Partnership, which has been co-funded by the European Commission.
**Conflicts of interest/Competing interests**-The authors declare that there is no conflict of interest.
**Ethics approval**-Not applicable
**Consent to participate**-Not applicable
**Consent for publication**-Not applicable
**Availability of data and material**-On request
**Authors' contributions**-A.B: Conceptualization, Methodology, Formal analysis, Investigation, Writing-Original Draft, Writing-Review and Editing, Visualization, Project administration. S.V.O: Conceptualization, Validation, Formal analysis, Investigation, Resources, Data Curation, Writing-Original Draft, Writing-Review and Editing, Visualization, Supervision.